\documentclass{article}
\usepackage{ijcai26}

\usepackage{times}
\usepackage{soul}
\usepackage{url}
\usepackage[hidelinks]{hyperref}
\usepackage[utf8]{inputenc}
\usepackage[small]{caption}
\usepackage{graphicx}
\usepackage{amsmath}
\usepackage{amsthm}
\usepackage{booktabs}
\usepackage{algorithm}
\usepackage{algorithmic}
\usepackage[switch]{lineno}

\usepackage{amsmath, amsfonts}
\usepackage{booktabs}
\usepackage{pifont} 
\newcommand{\cmark}{\ding{51}}  
\newcommand{\xmark}{\ding{55}}  

\usepackage{longtable}
\usepackage{lipsum}
\usepackage{booktabs}
\usepackage{multirow}

\title{Coarse-to-Fine Domain Incremental Learning with Attentive Distillation for Mining Footprint Segmentation in Multispectral Imagery}

\author{
Alif Tri Handoyo$^1$\thanks{Corresponding authors.}
\and
Vincent C.S. Lee$^{2}$
\and
Rizka Widyarini Purwanto$^{1}$
\and
Alex M. Lechner$^{1,3}$     
\and
Deanna Kemp$^{4}$\And
Muhamad Risqi U. Saputra$^{1}$\footnotemark[1]
\\
\affiliations
$^1$Monash University, Indonesia\\
$^2$Monash University, Australia\\
$^3$Northeastern University, China\\
$^4$The University of Queensland, Australia\\
\emails
\{alif.handoyo, vincent.cs.lee, rizka.purwanto, alex.lechner\}@monash.edu,
d.kemp@uq.edu.au,
risqi.saputra@monash.edu}

\begin{document}

\maketitle

\begin{abstract} Automatically mapping and segmenting global mining footprints using remote sensing and deep learning is critical for monitoring the socio-environmental risks and impacts of mining, yet its progress is hindered by the scarcity of fine-grained annotated data. Although large-scale datasets with coarse boundaries are widely available, leveraging them to improve fine-grained segmentation is challenging due to significant domain shift. To address this, we propose MineC2FNet, a coarse-to-fine domain incremental learning framework that exploits abundant coarse data to enhance fine-grained mining footprint segmentation. MineC2FNet adopts a teacher–student architecture with attentive distillation at both the feature and prediction levels, selectively transferring generalized knowledge from the coarse domain while enabling boundary refinement using limited fine-grained data (fine domain). We further introduce an expertly validated dataset of 219 images with precise boundary annotations across diverse geographies and commodities. Extensive experiments against state-of-the-art approaches, including domain adaptation and domain incremental learning methods, demonstrate that MineC2FNet achieves superior performance while effectively handling domain shift. The dataset and code are publicly available at \url{https://github.com/risqiutama/MineC2FNet}. \end{abstract}

\section{Introduction}

\paragraph{Mapping Global Mining Footprints: A Socio-Environmental Monitoring Challenge.} \hspace{1em}Mining is fundamental to the global economy, yet mineral resource extraction can cause profound, often irreversible damage to people and the environment \cite{Owen2023}. 

Extractive activities can lead to large-scale landscape alteration, ecosystem loss, biodiversity decline, and air and water pollution \cite{Sonter2018-er,Lechner2017}. Many of these impacts can displace local and downstream populations and adversely affect the health of local communities and indigenous peoples \cite{michelle2023,Owen2023}. These pressures and impacts can also drive resource-related conflicts and grievances. 

Despite the promise of local economic development, host communities often live with long-term environmental legacies, long after mines cease operations \cite{Kemp2025}. Multispectral remote sensing offers a method for continuous and repeatable tracking of mining footprints, in particularly in remote areas \cite{lechner2019historical,Maus2020,saputra2025multi}. This would enable scalable, remote monitoring of mining activity and its surrounding context, including how mining footprints change over time, and where change coincides with complex social and environmental risk factors. 

\begin{figure}
\centering
\includegraphics[width=0.42\textwidth]{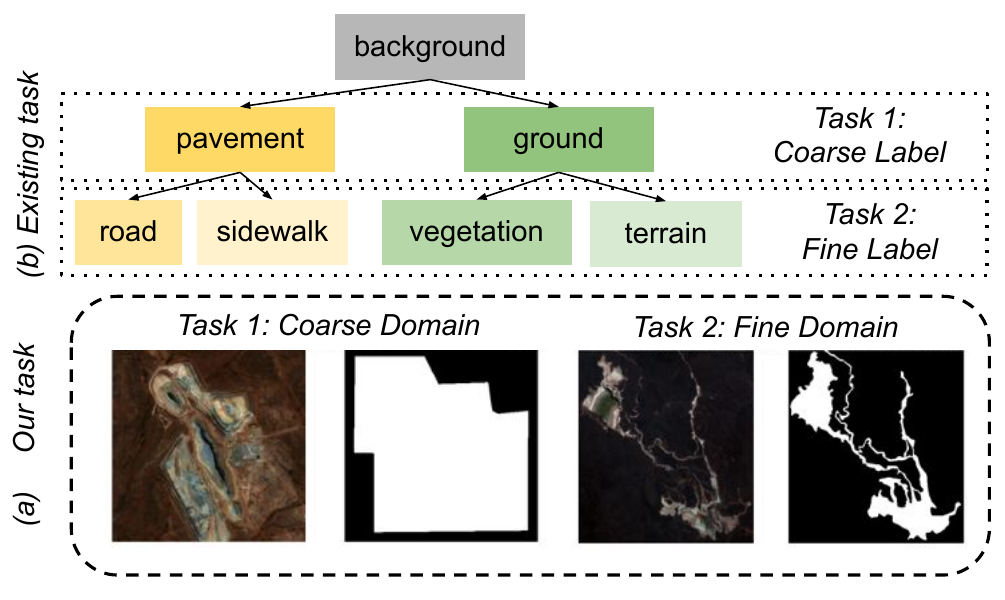}
\caption{Comparison between (a) our coarse-to-fine problem with (b) existing coarse-to-fine continual learning task.}
\label{fig:our_task}
\end{figure}

However, a critical bottleneck remains: while Deep Learning (DL)-based semantic segmentation offers a transformative alternative to labor-intensive manual mapping \cite{wieland2023semantic,cai2023learning,zhu2025}, these models depend on vast amounts of meticulously annotated data. Generating such data is exceptionally challenging, as distinguishing fine-grained features (e.g., pits, waste rock, tailings dams) requires substantial expert knowledge and field surveys \cite{Maus2020}. Although coarsely annotated data \cite{maus2022gmpv} offers a cost-effective alternative with extensive coverage, it merges distinct land cover features into imprecise footprints (Figure \ref{fig:our_task} (a) - Task 1). Models trained on such labels inherit these imprecisions, leading to inaccurate boundary delineation and hindering effective environmental monitoring. 

\paragraph{Our Contribution.} \hspace{1em}To leverage both abundant coarse dataset and scarce fine dataset for mining footprint segmentation, and to bridge their domain gap in incremental learning settings, we propose Mine Coarse-to-Fine Net (MineC2FNet). MineC2FNet uses \textit{attentive distillation} in a teacher–student training settings to adapt the model to new domains while selectively retaining knowledge from previous domains. The attentive distillation is applied at both feature (via \textit{attentive feature injection}) and prediction levels (via \textit{attentive knowledge transfer}) to transfer only high-quality delineation from the teacher (previous domain) to the student (new domain). Unlike prior coarse-to-fine continual learning methods defining "coarse" as broad label categories \cite{shenaj2022continual,Alfarra2024}, we introduce a new task where "coarse" refers to imprecise boundaries and "fine" to accurate, expert-validated ones. The contributions of this paper are:

\begin{itemize}
\item We present MineC2FNet, the first coarse-to-fine domain incremental learning framework for mining footprint segmentation using multispectral imagery.
\item We propose attentive distillation, a novel approach applying selective distillation to both feature and prediction levels.
\item We introduce a new, expertly validated dataset of 219 globally distributed images for fine-grained mining footprint segmentation.
\item We demonstrate the effectiveness of our framework and attentive distillation through comprehensive evaluation on this new dataset.
\end{itemize}

\paragraph{Team.} \hspace{1em}Our project brought together a multidisciplinary team in remote sensing, deep learning, and mining domain experts, including professionals from the Faculty of IT at Monash University, the College of Resources and Civil Engineering at Northeastern University, China, and the Centre for Social Responsibility in Mining at the University of Queensland. These experts were key to guiding the project conceptualisation, identifying detailed mining features, and validating the quality of our new dataset. More broadly, funded by the Ford Foundation, our research contributes to the foundations mission of advancing social and environmental justice by improving transparency and accountability in mining.
\section{Related Works}

\paragraph{Semantic Segmentation of Mining Footprint.} \hspace{1em}Applying deep learning-based semantic segmentation to multispectral satellite imagery has the potential to transform global mining footprint mapping and monitoring. Currently, global-scale studies, e.g., \cite{Owen2023}, rely on a single geographic coordinate to represent mining operations, failing to capture their true spatial extent. On the other hand, studies that leverage supervised semantic segmentation are typically mine- or region-specific \cite{qiao2024,tong2020land}, restricting their application to global datasets. Although \cite{saputra2025multi} has demonstrated mining segmentation on a global dataset, their model was trained on only 37 locations, limiting its generalization capability. Finally, the success of these methods is fundamentally dependent on the availability of large-scale, pixel-level annotated datasets, which are unfortunately not yet available.

\paragraph{Continual Learning in Remote Sensing.} \hspace{1em}Continual Learning (CL) addresses the challenge of training a model on a continuous stream of data or a sequence of tasks without experiencing \textit{catastrophic forgetting}, a condition in which the model tends to lose previously learned knowledge when trained on a new dataset or tasks. 

In remote sensing, the coarse-to-fine paradigm is applied to address CL problems in several distinct ways, ranging from data quality to architectural precision. To mitigate the problem of high annotation costs, some frameworks leverage coarse labels as a starting point to automatically generate more detailed, fine-grained pseudo-labels. These new pseudo labels then serve as the actual supervision to train the model. This strategy can be applied spatially, by turning imprecise polygons into sharp object boundaries \cite{Das2023}, or semantically, by using coarse labels (broad classes, e.g., cropland) to identify and classify fine labels (specific sub-classes, e.g., paddy field) \cite{Chen2024}. Similarly, \cite{shenaj2022continual} facilitates this coarse-to-fine CL by using knowledge distillation and a specialized weight initialization rule, allowing the model to learn new classes and adapt to new domains without being fully retrained. 

Beyond class-incremental shifts, domain-incremental learning (DIL) tackles geographical variations by adapting the models to new spectral and spatial distributions without requiring access to original training data. The GSMF-RS-DIL \cite{Huang2024transaction} framework utilizes graph space transformations and multifeature constraints to balance prior knowledge retention with the acquisition of new domain information. Similarly, multi-domain incremental architectures \cite{garg2022multi} employ domain-specific parameters and differential learning rates within Domain-Aware Residual Units to optimize the stability-plasticity trade-off. These integrated strategies allow remote sensing models to remain effective across diverse environments while significantly mitigating catastrophic forgetting.

\section{Data}

\paragraph{Coarse Dataset.} \hspace{1em}The coarse dataset was derived from the global-scale mining polygons (Version 2) data introduced by \cite{maus2022gmpv}, which originally contained over 44,929 mining polygons. This dataset provides a broader set of image features for mine footprint identification, characterized by rough, less precise edge delineations. These polygons are labeled as "Mining" and areas outside the polygons are considered "Non-Mining", resulting in a binary segmentation. To create a usable subset for our training, we filtered these polygons by establishing a minimum area threshold of approximately 9.8 km², corresponding to the smallest mine in our fine-grained dataset. Polygons with an area greater than this threshold were retained, resulting in a collection of 1,380 coarse-labeled images.  Satellite imagery for these locations were sourced from Sentinel-2 (10m) and Landsat 8 (15/30m) satellites and included  RGB, Near-Infrared (NIR), and Short-Wave Infrared (SWIR) bands.

\paragraph{Fine Dataset.} \hspace{1em}The fine dataset is a high-quality, expert-validated collection of images featuring precise and detailed boundaries for mining segmentation. This dataset consists of 219 images, which include RGB and NIR bands, with a subset of 200 images also containing SWIR bands. These data were manually harmonised from the original datasets of \cite{werner2020global,Lechner2016,lechner2019historical}, following methods outlined in \cite{saputra2025multi}. Similar to the coarse dataset, the resolution and extent of the multispectral data varied as we used Sentinel and Landsat 8. We split the dataset into training, validation, and test sets using an approximate 70:10:20 ratio. These locations was carefully balanced to account for a wide range of mine commodities, geographic regions, and climates. A key characteristic is its significant class imbalance, with non-mining pixels making up more than 80\% of the labels across the dataset.

\begin{figure}[t]
    \centering
    \includegraphics[width=\columnwidth]{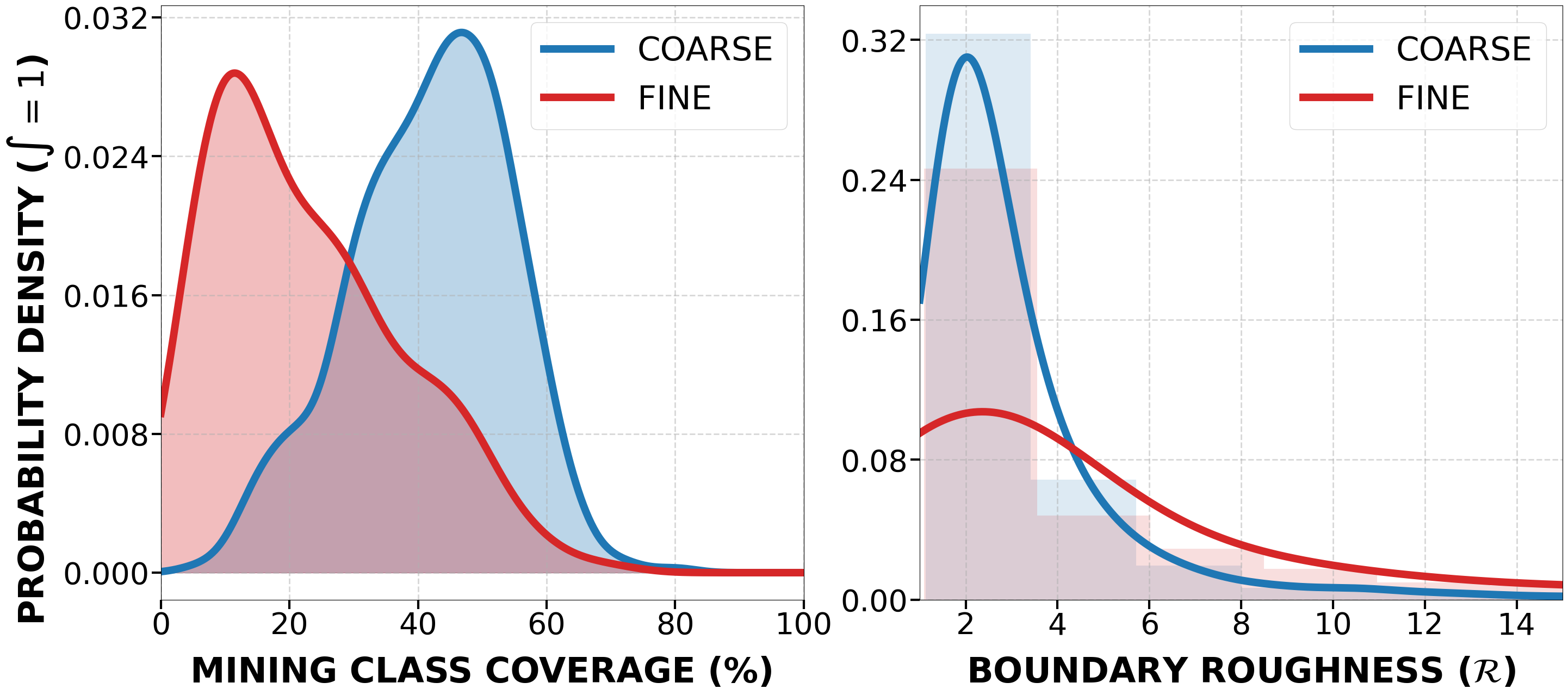}
    \caption{Domain shift analysis using Kernel Density Estimation (KDE). (Left) The coarse dataset exhibits a balanced class distribution, in contrast to the severe class imbalance of the fine dataset. (Right) Boundary roughness analysis demonstrating higher geometric complexity of fine masks compared to coarse masks.}
    \label{fig:domain_shift}
\end{figure}

\paragraph{Domain Shift.} \hspace{1em}Figure~\ref{fig:domain_shift} demonstrates a clear domain gap between the coarse and fine datasets. The mining class coverage analysis (left) reveals a significant disparity in class labels pixel distributions: Coarse annotations exhibit a relatively balanced occupancy peaking at $45\%$, whereas the fine dataset reflects the extreme sparsity of precise footprints, peaking at only $12\%$. The Boundary Roughness (right) measured as $\mathcal{R} = \frac{P^2}{4\pi A}$ further differentiates the masks roughness. Coarse masks peak at $\mathcal{R} \approx 2$, identifying them as smooth, simplified blobs. In contrast, fine labels capture the irregular, high-frequency boundary details of real-world mines, with values reaching up to $\mathcal{R} \approx 12$. Collectively, these metrics confirm that while coarse labels provide a simplified approximation, fine labels capture the true geometric complexity and natural class imbalance inherent in the target domain.

\begin{figure*}
    \centering
    \includegraphics[width=1\textwidth]{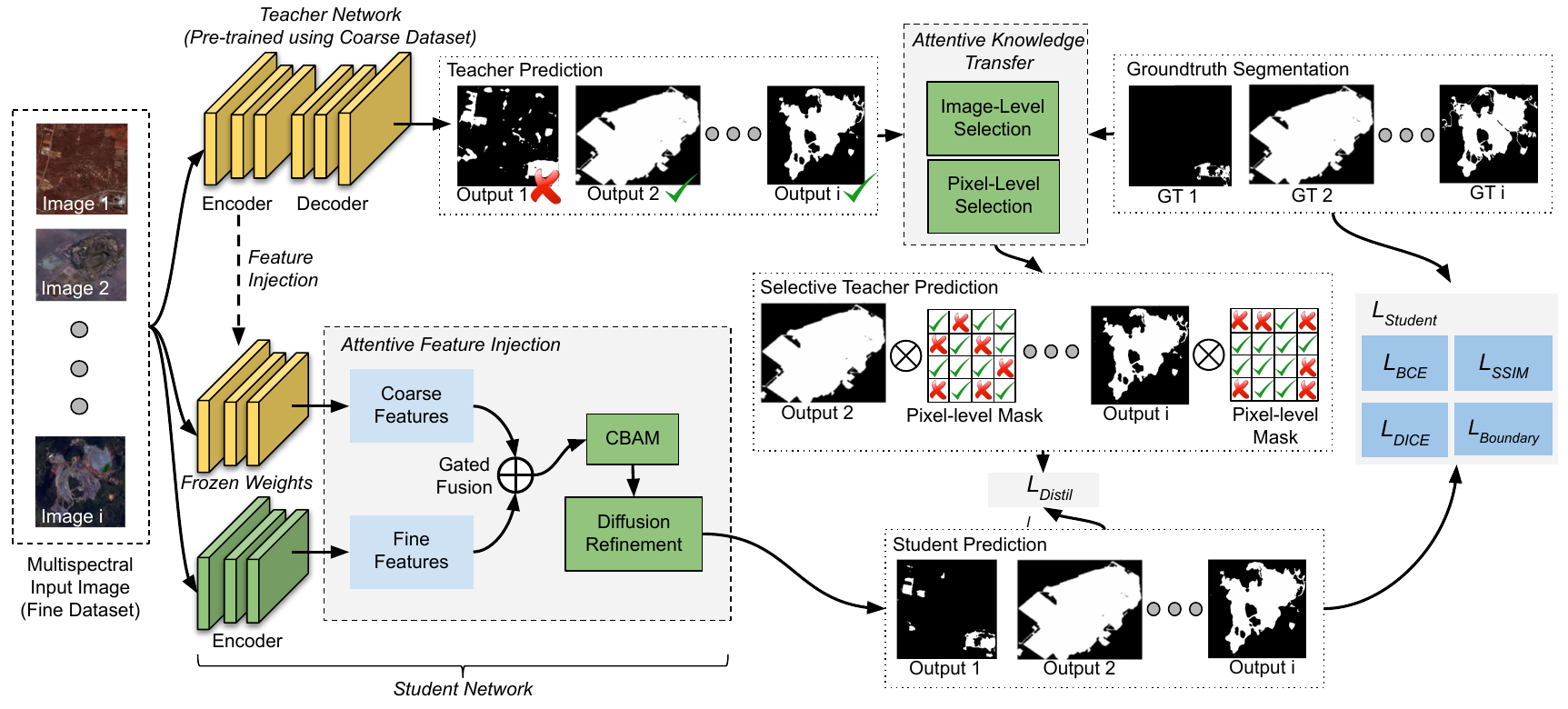}
    \caption{Overview of our coarse-to-fine domain incremental learning framework for mining footprint segmentation (MineC2FNet).}
    \label{fig:model_architecture}
\end{figure*}

\section{Methodology}
Our proposed method addresses the domain shift between coarse and fine datasets in training domain incremental learning model for mining footprint segmentation using multispectral satellite imagery. Figure \ref{fig:model_architecture} illustrates the overall workflow, highlighting our framework with attentive distillation applied at both the feature level (via Attentive Feature Injection) and the prediction level (via Attentive Knowledge Transfer). The core objective is to enable the model to learn from the new domain while selectively retaining knowledge from the previous domain. 
\subsection{Problem Formulation}
We frame our coarse-to-fine segmentation problem as a domain incremental learning problem consisting of two sequential tasks. The goal is to train a model $M$, parameterized by $\theta$, that leverages knowledge from an initial task trained on coarse domain $D_c$, to excel at a subsequent task with fine target domain $D_f$. This requires a domain incremental learning approach with selective attentive distillation, where the model is capable of transfering relevant general knowledge while discarding domain-specific noise (e.g., imprecise boundaries) from the coarse task.

Let $D_c = \{(X^{c}_i, Y^{c}_i)\}_{i=1}^{N_c}$ be the coarsely annotated dataset where $X^{c}_i$ is an input image, $Y^{c}_i$ is its corresponding coarse segmentation mask or label, and $N_c$ is the number of coarse training images. Let $D_f = \{(X^{f}_i, Y^{f}_i)\}_{i=1}^{N_f}$ be the fine-grained, expertly annotated dataset, where $Y^{f}_i$ is the precise, fine-grained mask, and $N_f$ is the total images in the fine dataset. Our model, $M(\theta)$, which consists of a teacher component $M(\theta_T)$ and a student component $M(\theta_S)$, where $\theta = \{\theta_T, \theta_S\}$, will be trained using the following steps:
\begin{enumerate}
    \item \textbf{Task 1 (Coarse Domain):} The teacher model $M(\theta_T)$ is trained on the coarse dataset $D_c$ to learn a set of generalized parameters $\theta_T$.
    \item \textbf{Task 2 (Fine Domain):} With the teacher $M(\theta_T)$ frozen, the student model $M(\theta_S)$ is trained on the fine dataset $D_f$ to learn its parameters $\theta_S$. During this phase, the student must learn to predict the fine labels $Y_f$ while selectively distilling knowledge from $M(\theta_T)$.
\end{enumerate}
The final objective is to optimize the parameters $\theta_S$ as described in Equation \ref{eq:problem_formulation}, where $L$ is the loss function. 

\begin{equation}
M(\theta_S) = \arg\min_{\theta_S} L(\theta_S | D_f, M(\theta_T))
\label{eq:problem_formulation}
\end{equation}

\subsection{Coarse-to-fine Domain Incremental Learning}
In the context of our coarse-to-fine domain incremental learning, the training for Task 1 follows the standard training procedure for semantic segmentation, utilizing Binary Cross-Entropy (BCE) \cite{mao2023} as the loss function. On the other hand, the training process for Task 2 is orchestrated by a composite loss function, $L_{\text{total}}$. This loss, shown in Eq. (\ref{eq:total_loss}), is the core of our domain incremental learning approach. It combines $L_{\text{student}}$, the student's loss function on the fine data, with $L_{\text{distill}}$, the attentive distillation loss from the teacher.
\begin{equation}
L_{\text{total}}(D_f) = L_{\text{student}} + L_{\text{distill}}
\label{eq:total_loss}
\end{equation}

\paragraph{Student Loss.} \hspace{1em}The student loss, $L_{\text{student}}$, is a comprehensive objective function designed to train the student model $M(\theta_S)$ on the fine-grained dataset $D_f$. As seen in Eq. (\ref{eq:student_loss}), it consists of four components, including Binary Cross-Entropy ($L_{\text{BCE}}$), Dice Loss ($L_{\text{Dice}}$), Structural Similarity Loss ($L_{\text{SSIM}}$), and Boundary Loss ($L_{\text{Boundary}}$) functions. 
\begin{equation}
L_{\text{student}}(D_f) = L_{\text{BCE}} + L_{\text{Dice}} + L_{\text{SSIM}} + L_{\text{Boundary}}
\label{eq:student_loss}
\end{equation}
$L_{\text{BCE}}$ is utilized as a widely used loss function for binary classification tasks, which we apply on a pixel-wise basis to measure the difference between the student's prediction $\hat{Y}^{f}_i$ and the ground truth fine mask $Y^{f}_i$. The Dice loss is added since it is particularly effective for segmentation tasks with imbalanced classes by maximizing the overlap between $\hat{Y}^{f}_i$ and $Y^{f}_i$ \cite{Sudre2017}. Furthermore, we also incorporate $L_{SSIM}$ \cite{zhao2019} as this loss encourages the model to preserve the structural information of $Y^{f}_i$, which is crucial for maintaining realistic mining shapes.

Besides the loss functions above, we also incorporate $L_{\text{Boundary}}$ to explicitly improve the model performance on edge delineation. This loss is designed by generating a boundary importance map from the ground truth mask and then using it to weight the $L_{\text{BCE}}$. In particular, we first apply Sobel filters to $Y^{f}_i$ to compute the horizontal ($G_x$) and vertical ($G_y$) gradients. The boundary importance map $\hat{W} \in [0,1]$ is then calculated as in Eq. (\ref{eq:boundary_norm}).
\begin{equation}
\hat{W} = \frac{W - W_{\min}}
{W_{\max} - W_{\min}}, \text{where} \, W = \sqrt{G_x^2 + G_y^2}
\label{eq:boundary_norm}
\end{equation}
$L_{\text{Boundary}}$ is then defined as a weighted binary-cross entropy loss as described in Eq. (\ref{eq:boundary_loss}), where the weights are derived from image edges to emphasize optimizing the boundary regions of the segmented images.
\begin{equation}
L_{\text{Boundary}} = \frac{1}{K} \sum_{j=1}^{K} \hat{W}_j \cdot L_{\text{BCE}_j}
\label{eq:boundary_loss}
\end{equation}
Note that $K$ represents the number of pixel in the images. This formulation forces the model to pay more attention to correctly classifying pixels on the segmented boundaries.

\paragraph{Distillation Loss.} \hspace{1em}We adopt a knowledge distillation where the loss ($L_{\text{Distill}}$) is formulated as the standard Kullback-Leibler (KL) divergence between the softened probability distributions of the teacher ($P^T$) and student ($Q^T$), scaled by a temperature factor $T$. As the teacher is trained in a domain with imperfect coarse labels, we introduce attentive distillation to ensure only reliable knowledge is transferred. This mechanism is applied at both the feature and prediction levels, as we will detail next.

\subsection{Attentive Feature Injection}
Our attentive feature injection is a multi-stage pipeline designed to intelligently transfer generalized feature extraction capabilities of the teacher model $M(\theta_T)$ to the student $M(\theta_S)$. This pipeline was designed to solve our domain incremental challenge with the understanding that $M(\theta_T)$ was trained on a larger set (source domain), capturing a more visually diverse range of global mining footprints. As a result, $M(\theta_T)$ possesses generalized knowledge of how mining areas appear in satellite imagery, even though its boundary delineation remains imprecise. To this end, rather than simply fine-tuning $M(\theta_T)$'s weight in the Task 2 or even concatenating $M(\theta_T)$'s and $M(\theta_S)$'s features, our attentive feature injection selectively fuses, refines, and enhances feature maps from both the teacher (generalized source domain features that uses coarse data) and student (specific target domain features that uses fine data). As illustrated in Figure \ref{fig:model_architecture}, the pipeline consists of four stages: Feature Injection, Gated Fusion (GF), Convolutional Block Attention Module (CBAM), and Diffusion Refinement (DR).

\paragraph{Feature Injection.} \hspace{1em}A key distinction in our attentive distillation approach is that we inject a subset of $M(\theta_T)$, particularly the feature extractor, to $M(\theta_S)$, enabling $M(\theta_S)$ to incorporate dual, identical feature extractors: a "teacher feature extractor" (a frozen copy of the teacher's backbone, part of $\theta_T$) and a "student feature extractor" (with trainable parameters, part of $\theta_S$). During Task 2, for a given input image $X^{f}_i$, these two backbones operate in parallel. The student backbone extracts high-fidelity features directly from the fine-grained target domain input, while the frozen teacher backbone provides a parallel stream of generalized feature maps derived from the source domain. These two sets of features, one rich in precise target domain detail ($F_{\text{fine}}$) and the other in broad source domain context ($F_{\text{coarse}}$), are then passed to the subsequent attentive fusion and refinement stages.

\paragraph{Gated Fusion (GF).} \hspace{1em}To dynamically merge contextual information from the coarse teacher backbone with high-frequency details from the fine student backbone, we employ Gated Fusion \cite{takikawa2019}. This module’s merit lies in its learnable, pixel-wise gating mechanism, which adaptively decides the contribution of each feature source. This allows the network to prioritize general context or precise detail as needed, creating a richer, fused representation without the need for fixed fusion rules.

\paragraph{CBAM.} \hspace{1em}The fused features are then refined using the Convolutional Block Attention Module \cite{Sanghyun2018}. We use CBAM for its efficiency and effectiveness in feature enhancement. Its primary benefit is that it sequentially infers attention maps along both channel and spatial dimensions. This process allows the model to learn what features are most salient (channel-wise) and where to focus its attention (spatial-wise), further boosting the discriminative power of the representations.

\paragraph{Diffusion Refinement (DR).} \hspace{1em}As a final stage, we introduce a new Diffusion Refinement module to produce a highly polished representation for segmentation. Inspired by iterative refinement frameworks \cite{Guosheng2017}, our module progressively enhances features by propagating information through a sequence of custom refinement blocks. The key idea of our design is the structure of each block, which combines a convolution, a CBAM layer, and a residual connection to effectively sharpen noisy features. For each refinement step $t \in [1, T]$, the feature map $F_{t-1}$ is updated using the operations detailed in Eq. (\ref{eq:conv_temporal})-(\ref{eq:residual_refine}).
\begin{equation}
F_{\text{conv}_t} = \text{Conv}_{3 \times 3}(F_{t-1})
\label{eq:conv_temporal}
\end{equation}
\begin{equation}
F_{\text{att}_t} = \text{CBAM}(F_{\text{conv}_t})
\label{eq:cbam_temporal}
\end{equation}
\begin{equation}
F_t = F_{\text{att}_t} + F_{t-1}
\label{eq:residual_refine}
\end{equation}
This iterative process results in a final feature map, $F_{refined}$, with an enhanced focus on the most salient regions and channels, making it highly effective for achieving accurate semantic segmentation.

\subsection{Attentive Knowledge Transfer}
In addition to attentive feature injection, we introduce an attentive knowledge transfer mechanism at the prediction level. A key challenge is that a standard teacher trained on the coarse domain is not infallible; unconditional (non-selective) distillation transfers domain-specific noise like imprecise boundaries, degrading performance. To mitigate this, our framework adaptively selects \textit{when} and \textit{where} to apply the distillation loss, ensuring that the student only learns from the teacher's most confident and accurate predictions. 

\paragraph{Image-level Selection.} \hspace{1em}The first layer of our attentive mechanism operates at the image level, aiming to identify images within a training batch where the teacher model $M(\theta_T)$ produces more accurate predictions than the student model $M(\theta_S)$. This is based on the hypothesis that distillation is beneficial only when $M(\theta_T)$ is more accurate than $M(\theta_S)$ for a given input image. We formalize this by utilizing an indicator function $\mathbb{I}(\cdot)^{\text{img}}$ which returns 1 if $L_{\text{BCE}}(Y^{f}_i, P_i) < L_{\text{BCE}}(Y^{f}_i, Q_i)$ and 0 otherwise. Note that $P_i$ and $Q_i$ denote the output prediction (logits) of $M(\theta_T)$ and $M(\theta_S)$, respectively, for image $i$. The distillation loss with image-level selection is then formulated in Eq. (\ref{eq:image_level_distill}).
\begin{align}
L_{\text{distill}}^{\text{image}} = \frac{T^2}{N_f} \sum_{i=1}^{N_f}
\mathbb{I}\Bigg(
\begin{aligned}
&L_{\text{BCE}}(Y^{f}_i, P_i) \\
&< L_{\text{BCE}}(Y^{f}_i, Q_i)
\end{aligned}
\Bigg)^{\text{img}} \cdot \text{KL}(P^T_i \parallel Q^T_i)
\label{eq:image_level_distill}
\end{align}

\paragraph{Pixel-level Selection.} \hspace{1em}The second layer of attentive distillation applies at the pixel level. In the context of semantic segmentation, as $L_{\text{distill}}$ will be computed for every single pixel, we realize that not all per-pixel teacher's prediction will be accurate, especially when the teacher was trained using imprecise ground truth labels. To this end, we selectively apply distillation only when the teacher's prediction at the pixel level is more accurate than the student's prediction. Similar to image-level selection, we formalize this pixel-level selection by utilizing an indicator function $\mathbb{I}(\cdot)^{\text{pxl}}$ which returns 1 if $L_{\text{BCE}}(Y^{f}_{ij}, P_{ij}) < L_{\text{BCE}}(Y^{f}_{ij}, Q_{ij})$ and 0 otherwise. Note that $j \in K$ represents the pixel index in image $i$. The distillation loss with pixel-level selection is formulated in Equation \ref{eq:pixel_level_distill}.
\begin{align}
L_{\text{distill}}^{\text{pixel}} = \frac{T^2}{N_f} \sum_{i,j}
\mathbb{I}\Bigg(
\begin{aligned}
&L_{\text{BCE}}(Y^{f}_{ij}, P_{ij}) \\
&< L_{\text{BCE}}(Y^{f}_{ij}, Q_{ij})
\end{aligned}
\Bigg)^{\text{pxl}} \cdot \text{KL}(P^T_{ij} \parallel Q^T_{ij})
\label{eq:pixel_level_distill}
\end{align}
Note that in practice, $\mathbb{I}^{\text{pxl}}$ will be in a matrix form rather than in a scalar form as in $\mathbb{I}^{\text{img}}$.
\paragraph{Hybrid Selection.} \hspace{1em}Our final approach combines both image-level and pixel-level selection 
strategies into a hybrid selection mask $\mathbb{I}^{\text{hybrid}}$ by performing element-wise multiplication of $\mathbb{I}^{\text{pxl}}$ and $\mathbb{I}^{\text{img}}$ as shown in Eq. (\ref{eq:hybrid_mask_simple}).

\begin{equation}
\mathbb{I}^{\text{hybrid}}_{ij} = \mathbb{I}^{\text{img}}_i \cdot \mathbb{I}^{\text{pxl}}_{ij}
\label{eq:hybrid_mask_simple}
\end{equation}
This ensures that distillation is applied only to pixels where the teacher is more accurate than the student, and only within input images where the teacher demonstrates overall superior performance. The final distillation loss is computed as defined in Equ. (\ref{eq:masked_distill}):
\begin{equation}
L_{\text{distill}} = \frac{\sum_{i,j} \left( \mathbb{I}^{\text{hybrid}}_{ij} \cdot T^2 \cdot \text{KL}(P^T_{ij} \parallel Q^T_{ij}) \right)}{\sum_{i,j} \mathbb{I}^{\text{hybrid}}_{ij} + \epsilon}
\label{eq:masked_distill}
\end{equation}
where the numerator accumulates the KL divergence values, and the denominator normalizes it via the total number of selected pixels (with $\epsilon$ for numerical stability). This hierarchical and attentive strategy ensures that the student selectively learns only from the teacher’s most reliable predictions.

\subsection{Training}
\paragraph{Experimentation Setting.} \hspace{1em}We implemented our framework using a Feature Pyramid Network (FPN) with a DenseNet-121 backbone, initialized with ImageNet pre-trained weights and modified to process 6-channel inputs (RGB, NIR, SWIR) at a resolution of $256 \times 256$. 
The model was trained using the AdamW optimizer \cite{loshchilov2019} with an initial learning rate of $1 \times 10^{-3}$, weight decay of $1 \times 10^{-4}$, and a batch size of 8 for a maximum of 100 epochs. Standard geometric augmentations, including random flips, rotations, and scaling, were applied to enhance generalization. All experiments were conducted on an NVIDIA RTX A4000 GPU.

\paragraph{Evaluation Metrics and Baselines.} \hspace{1em}To assess performance, we benchmark MineC2FNet against five distinct learning paradigms: Transfer Learning (including the foundation model, e.g., Prithvi), Domain Adaptation, Baseline Continual Learning, Class Incremental Learning (CIL), and Domain Incremental Learning (DIL). Performance is evaluated using three standard metrics: \textit{Pixel Accuracy}, \textit{Mean F1 Score (mF1)}, and \textit{Mean Intersection over Union (mIoU)}. More details can be found in the supplementary material.

\begin{table}[t]
\centering
\small 
\setlength{\tabcolsep}{0pt}
\begin{tabular*}{\columnwidth}{@{\extracolsep{\fill}}p{0.58\columnwidth}p{0.14\columnwidth}p{0.14\columnwidth}p{0.14\columnwidth}@{}}
\toprule
\textbf{Model} & \textbf{Acc.} & \textbf{mF1} & \textbf{mIoU} \\
\midrule
\multicolumn{4}{@{}l}{\textbf{\textit{CNN and Transformer-based methods (Transfer Learning)}}} \\
U-Net \cite{ronneberger2015}        & 89.79 & 79.52 & 67.26 \\
DeepLabV3+ \cite{chen2018}          & 90.01 & 79.57 & 67.47 \\
FPN \cite{Lin2017}                  & 90.65 & 80.06 & 68.26 \\
Prithvi \cite{jakubik2023}          & 66.14 & 66.21 & 49.74 \\
\midrule
\multicolumn{4}{@{}l}{\textbf{\textit{Domain Adaptation}}} \\
UDAforRS \cite{Li2022remotesensing}         & 75.16 & 72.42 & 57.49 \\
BUS \cite{choe2024open}                  & 60.22 & 59.21 & 42.35 \\
\midrule
\multicolumn{4}{@{}l}{\textbf{\textit{Baseline Continual Learning}}} \\
LwF \cite{Zhizhong2018}             & 90.22 & 78.94 & 67.02 \\
LwM \cite{dhar2019}                 & 90.60 & 79.36 & 67.43 \\
Replay \cite{rolnick2019}           & 88.81 & 76.50 & 63.61 \\
\midrule
\multicolumn{4}{@{}l}{\textbf{\textit{Class Incremental Learning}}} \\
SPPA \cite{Lin2022computervision}                & 54.68 & 48.32 & 31.85 \\
CCDA \cite{shenaj2022continual}                & 78.90 & 72.90 & 57.00 \\
\midrule
\multicolumn{4}{@{}l}{\textbf{\textit{Domain Incremental Learning}}} \\
MDIL-SS \cite{garg2022multi}                & 88.62 & 85.36 & 71.99 \\
GSMF-RS-DIL \cite{Huang2024transaction}          & 88.83 & 85.02 & 71.25 \\
\midrule
\textbf{MineC2FNet (ours)}          & \textbf{92.33} & \textbf{84.10} & \textbf{73.64} \\
\bottomrule
\end{tabular*}
\caption{Performance comparison of different models across various models and learning paradigms.}
\label{tab:model_comparison}
\end{table}

\begin{figure}[t]
   \centering
   \includegraphics[width=1\columnwidth]{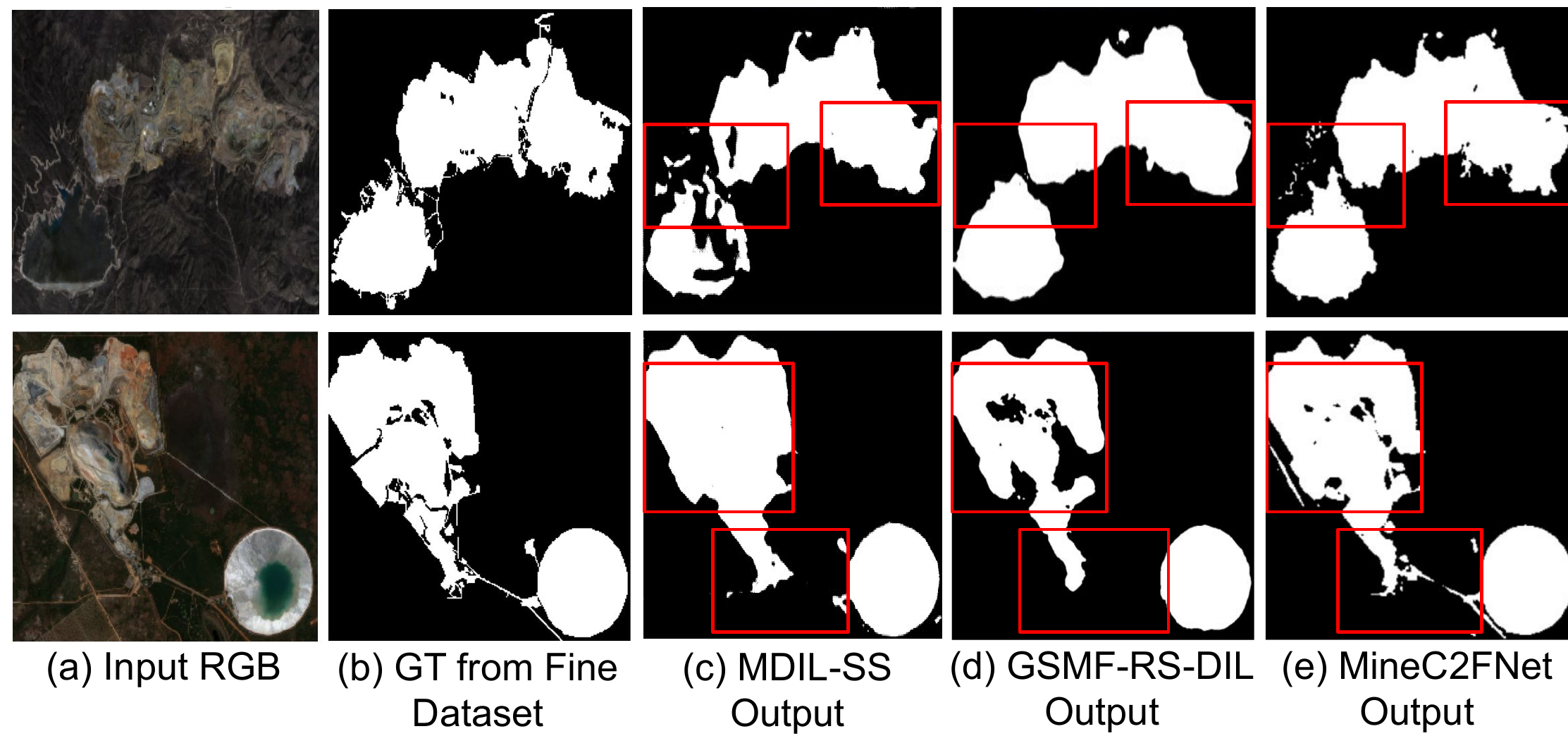} 
   \caption{Segmentation results of MineC2FNet and several baseline models on the fine-grained test set.}
   \label{fig:results_comparison}
\end{figure}

\begin{figure}[t]
    \centering
    \includegraphics[width=1\columnwidth]{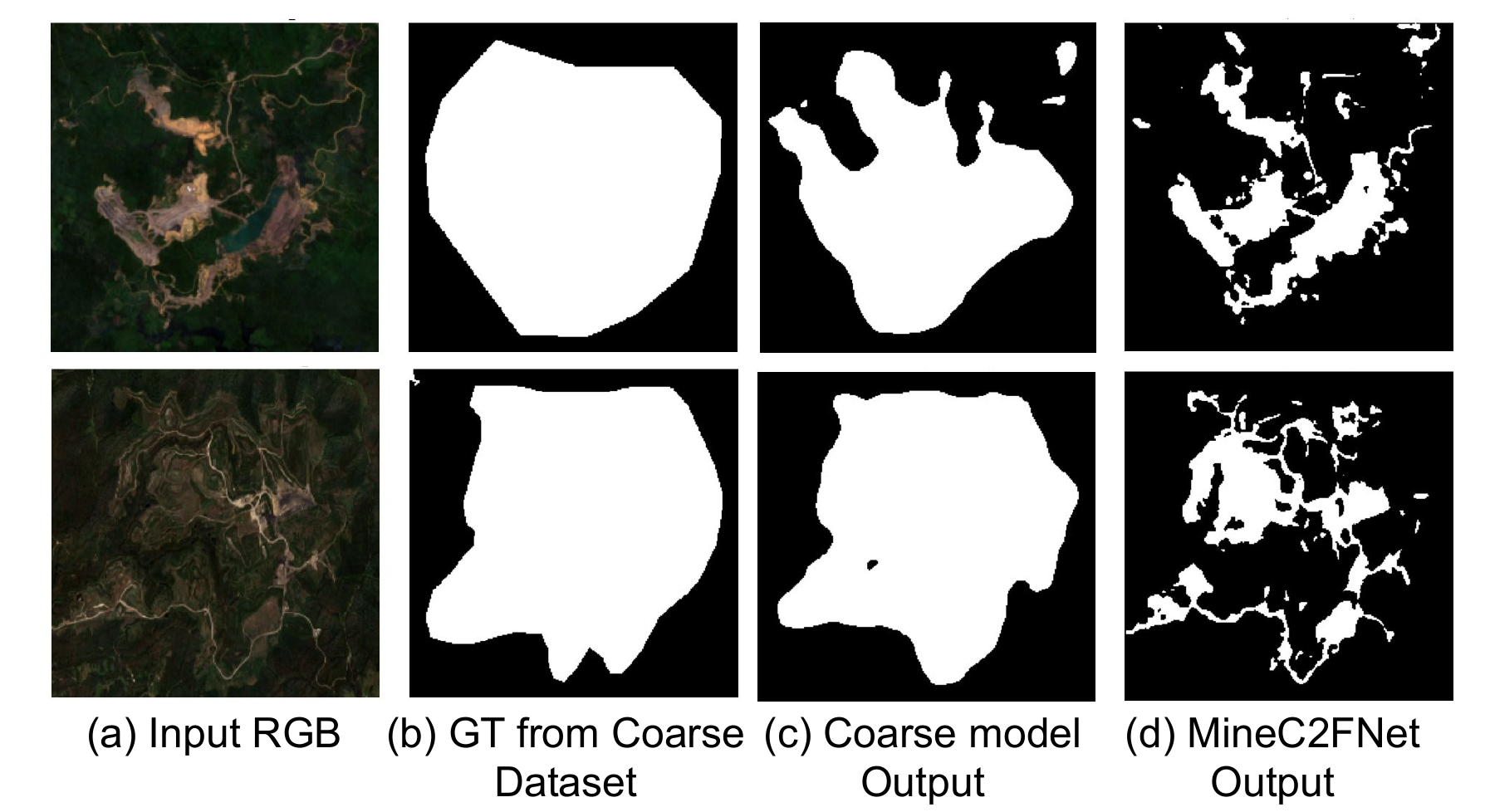}
    \caption{Through our framework, MineC2FNet produces more fine-grained output in the coarse dataset.}
    \label{fig:coarse_prediction}
\end{figure}

\section{Results and Discussion}

\subsection{Overall Performance}
\paragraph{Quantitative Analysis.} \hspace{1em}Table \ref{tab:model_comparison} summarizes the performance of MineC2FNet against state-of-the-art methods. Our model, termed MineC2FNet, demonstrates superior performance, achieving the highest Accuracy (92.33\%) and mIoU (73.64\%) compared to different methods across five distinct learning categories. It is worth noting that while leading DIL methods like MDIL-SS achieve a slightly higher mF1 score (85.36\% vs. 84.10\%), our framework secures a clear advantage in mIoU, surpassing them by over 1.6 percentage points. This numerical superiority translates directly to visual precision, as shown in Figure \ref{fig:results_comparison}, in which MineC2FNet successfully captures more detailed boundaries compared to the existing DIL models. Conversely, specialized CIL and Domain Adaptation methods struggle with this specific coarse-to-fine domain shift. Notably, methods such as SPPA (31.85\%) and BUS (42.35\%) perform significantly worse than even standard baseline continual learning approaches like LwF or LwM. This confirms that standard class-focused or domain adaptation strategies are ill-suited for this problem, whereas our approach successfully produces fine details while preserving coarse general knowledge.

We further compare our method with the coarse baseline on the coarse-grained dataset, as reported in Table \ref{tab:coarse_comparison}. Although MineC2FNet achieves a lower mIoU (44.29\%) than the coarse baseline (72.49\%), this outcome is expected due to the imprecise annotations in the coarse dataset. The coarse baseline is trained and evaluated entirely on the source domain, and its predictions therefore closely match the coarse labels, which are known to contain inaccurate boundaries. In contrast, our model adapts to the target domain with finer and more precise annotations (see Figure \ref{fig:coarse_prediction}), resulting in lower agreement when evaluated against coarse labels. This discrepancy does not indicate a failure of our approach; rather, it reflects superior boundary delineation while maintaining generalization by retaining useful knowledge from both coarse and fine domain. Evaluations across five Köppen–Geiger climate zones (Beck et al. 2018) confirmed the model's global robustness and generalizability (see supplementary material).

\begin{table}[t]
\centering
\small
\setlength{\tabcolsep}{5pt} 
\begin{tabular}{lcc}
\toprule
\textbf{Metric} & \textbf{Coarse Baseline} & \textbf{MineC2FNet (Ours)} \\
\midrule
Accuracy & 0.8733 & 0.7681 \\
mF1      & 0.8331 & 0.5612 \\
mIoU     & 0.7249 & 0.4429 \\
\bottomrule
\end{tabular}
\caption{Performance comparison between the coarse baseline and the proposed MineC2FNet.}
\label{tab:coarse_comparison}
\end{table}

\paragraph{Qualitative Analysis.} \hspace{1em}Figure \ref{fig:results_comparison} highlights the precise segmentation masks generated by MineC2FNet, demonstrating its ability to accurately delineate complex boundaries. As shown in the second row, our model reduces false positives caused by noisy features more effectively than the baselines. Furthermore, when tested on the coarse dataset, our domain-adapted MineC2FNet yields fine-grained segmentation that reflects true features on the ground (Figure \ref{fig:coarse_prediction}(d)). In contrast, training the model only on the coarse dataset inherits imprecise identification and poor boundary delineation (Figure \ref{fig:coarse_prediction}(c)), underscoring the importance of our coarse-to-fine domain incremental learning.

\paragraph{Model Complexity and Efficiency.} \hspace{1em}Finally, we evaluate MineC2FNet efficiency against the FPN baseline. The model achieves a substantial 5.4 point mIoU improvement, with a negligible speed reduction (13.38 to 11.00 FPS) due to added attention mechanisms. We consider this trade-off justified by the performance gain (detailed results in supplementary material).

\begin{table}[t]
\centering
\small
\setlength{\tabcolsep}{1mm}
\begin{tabular}{ccc|cccc}
\toprule
 \textbf{GF} & \textbf{CBAM} & \textbf{DR} & 
\textbf{Acc.} & \textbf{mF1} & \textbf{mIoU} & $\Delta$ \textbf{mIoU} \\
\midrule
 \xmark & \xmark & \xmark & 89.00 & 75.39 & 62.35 & -- \\
 \cmark & \xmark & \xmark & 90.31 & 76.72 & 64.35 & +2.00 \\
 \cmark & \cmark & \xmark & \textbf{91.33} & 79.15 & 67.79 & +5.44 \\
 \cmark & \cmark & \cmark & 91.08 & \textbf{79.84} & \textbf{68.41} & \textbf{+6.06} \\
\bottomrule
\end{tabular}
\caption{Ablation study of Attentive Feature Injection.}
\label{tab:impact_of_feature_extraction}
\end{table}

\begin{table}[t]
\centering
\small
\begin{tabular}{lcccc}
\toprule
\textbf{Strategy} & \textbf{Acc.} & \textbf{mF1} & \textbf{mIoU} & $\Delta$ \textbf{mIoU} \\
\midrule
No Distill. (Base)    & 89.00  & 75.39 & 62.35 & -- \\
Standard Distill.     & 89.38  & 75.79 & 62.59 & +0.24 \\
Image-level Sel.      & 90.42  & 78.79 & 67.17 & +4.82 \\
Pixel-level Sel.      & \textbf{91.14}  & 79.30 & 67.84 & +5.49 \\
Hybrid Sel.           & 91.10  & \textbf{79.61} & \textbf{67.91} & \textbf{+5.56} \\
\bottomrule
\end{tabular}
\caption{Comparison of different Attentive Knowledge Transfer strategies (all configurations include feature injection).}
\label{tab:distill_strategies}
\end{table}

\begin{table}[t]
\centering
\small
\begin{tabular}{lcccc}
\toprule
\textbf{Band Comb.} & \textbf{Acc.} & \textbf{mF1} & \textbf{mIoU} & $\Delta$ \textbf{mIoU} \\
\midrule
RGB (Baseline)          & 87.08 & 71.12 & 57.09 & -- \\
RGB+NIR               & 89.00 & 75.39 & 62.35 & +5.26 \\
RGB+NIR+SWIR        & \textbf{89.68} & \textbf{79.29} & \textbf{67.29} & \textbf{+10.20} \\
\bottomrule
\end{tabular}
\caption{Impact of different band combinations.}
\label{tab:band_comparison}
\end{table}

\subsection{Ablation Study}
\paragraph{Impact of Attentive Feature Injection.} \hspace{1em}We conducted ablation studies to clearly isolate the effects of each module within our attentive feature injection mechanism, i.e., Gated Fusion (GF), CBAM, and Diffusion Refinement (DR). Note that only RGB and NIR bands were used for this experiment. As shown in Table \ref{tab:impact_of_feature_extraction}, while adding GF alone can improves performance, incorporating CBAM and DR yields a significant performance gain. The highest mIoU (68.41\%) is achieved when all three components are combined. This confirms that these modules contribute critically to the complete pipeline for optimal fusion of coarse and fine features.

\paragraph{Impact of Attentive Knowledge Transfer.} \hspace{1em}Table \ref{tab:distill_strategies} highlights the impact of different distillation methods. A model trained with standard distillation shows only marginal improvement over the baseline, confirming the risk of negative knowledge transfer. In contrast, our attentive distillation, at either image or pixel level, shows a large improvement ($\geq 4.82$ $\Delta$mIoU). The hybrid method, distilling knowledge only from the teacher's most accurate regions, achieves the highest performance, validating that a selective, hierarchical mechanism is critical for successful coarse-to-fine domain knowledge transfer.

\paragraph{Bands Comparison.} \hspace{1em}As shown in Table \ref{tab:band_comparison}, incorporating non-visible spectral bands is critical. While adding NIR significantly increases performance, the optimal results are achieved using the full RGB-NIR-SWIR combination. This underscores the value of NIR and SWIR bands in identifying distinct mining footprint features \cite{saputra2025multi}.

\section{Conclusion} In this paper, we addressed the critical challenge of domain shift between coarse (abundant) and fine (scarce) datasets in training domain incremental learning model for mining footprint segmentation. We introduce MineC2FNet, a novel coarse-to-fine domain incremental framework. Our core contribution, an attentive distillation mechanism, proved highly effective, achieving a state-of-the-art mIoU of 73.64\% by successfully selectively transferring knowledge from coarse to fine-grained data and outperforming other continual learning models. This work also introduced a new, expertly validated global dataset to facilitate further research in this domain. However, given our current limitation to binary segmentation, future work will extend this framework to complex multi-class segmentation of diverse mining and non-mining categories.

\section*{Acknowledgments}
This study was supported in part by the Google Research Scholar Program, Australian Research Council LP200301160, and the Ford Foundation. The work of A. T. H. was partially supported by the Monash University Indonesia PhD Tuition Scholarship. We also sincerely thank John R. Owen for his contribution to the development of the training data.

\bibliographystyle{named}
\bibliography{ijcai26}

\clearpage
\section{Supplementary Material}\

\onecolumn

\begin{figure}[t]
   \centering
   \includegraphics[width=0.99\columnwidth]{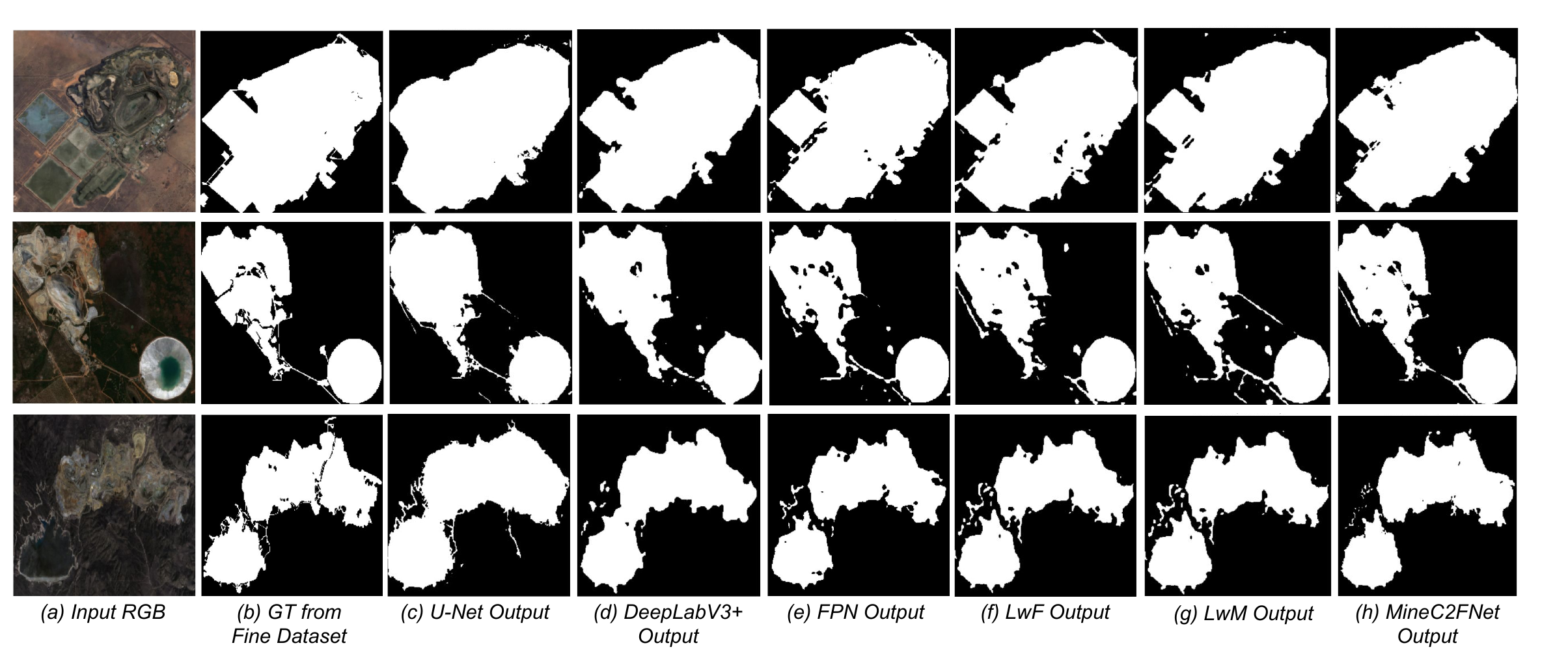} 
   \caption{Segmentation results of MineC2FNet and several baseline models on the fine-grained test set.}
   \label{fig:results_comparison}
\end{figure}

\begin{center}
\setlength{\LTcapwidth}{\textwidth}
\begin{longtable}{@{}lllllll@{}}
\label{tab:training_set} \\
\toprule
\textbf{Mine ID} & \textbf{Mine Name} & \textbf{Country} & \textbf{Commodity} & \textbf{Climate} & \textbf{Region} & \textbf{Sensor} \\
\midrule
\endfirsthead

\multicolumn{7}{c}{{}} \\
\toprule
\textbf{Mine ID} & \textbf{Mine Name} & \textbf{Country} & \textbf{Commodity} & \textbf{Climate} & \textbf{Region} & \textbf{Sensor} \\
\midrule
\endhead

\bottomrule
\caption{Details of the Training Set (153 images) (Continued on next page).} 
\endfoot

\bottomrule
\caption{Details of the Training Set (153 images).} 
\endlastfoot

3 & Alpamarca & Peru & PbZnCu & 5 & South America & Sentinel 2 \\
5 & Atacocha & Peru & PbZnCu & 5 & South America & Sentinel 2 \\
14 & Carmen\_de\_Andacollo & Chile & Cu & 2 & South America & Sentinel 2 \\
17 & Cerro\_de\_Pasco & Peru & PbZnCu & 5 & South America & Sentinel 2 \\
18 & Cerro\_Lindo & Peru & PbZnCu & 2 & South America & Sentinel 2 \\
29 & Flying\_Fox\_Spotted\_Quoll & Australia & Ni S & 2 & Oceania & Sentinel 2 \\
31 & Impala\_Group & South Africa & PGEs & 2 & Africa & Sentinel 2 \\
32 & Julcani & Peru & PbZnCu & 5 & South America & Sentinel 2 \\
34 & Kimberley & South Africa & Diamonds & 2 & Africa & Sentinel 2 \\
37 & Letseng & Lesotho & Diamonds & 5 & Africa & Sentinel 2 \\
38 & Magellan & Australia & PbZn & 2 & Oceania & Sentinel 2 \\
39 & Mantoverde & Chile & Cu & 2 & South America & Sentinel 2 \\
41 & McArthur\_River\_Australia & Australia & PbZn & 2 & Oceania & Sentinel 2 \\
43 & Mogalakwena & South Africa & PGEs & 2 & Africa & Sentinel 2 \\
44 & Mototolo & South Africa & PGEs & 3 & Africa & Sentinel 2 \\
46 & Nkomati & South Africa & Ni S & 3 & Africa & Sentinel 2 \\
47 & North\_Mara & Tanzania & Au & 1 & Africa & Sentinel 2 \\
50 & Paracatu & Brazil & Au & 1 & South America & Sentinel 2 \\
51 & Pinto\_Valley & USA & CuMo & 2 & North America & Sentinel 2 \\
55 & Punitaqui & Chile & CuAg & 2 & South America & Sentinel 2 \\
58 & Stawell & Australia & Au & 3 & Oceania & Sentinel 2 \\
60 & Tara & Ireland & PbZn & 3 & Europe & Sentinel 2 \\
62 & Tizapa & Mexico & PbZnCu & 1 & North America & Sentinel 2 \\
64 & Two\_Rivers & South Africa & PGEs & 3 & Africa & Sentinel 2 \\
65 & Union\_North\_South & South Africa & PGEs & 2 & Africa & Sentinel 2 \\
66 & Velardena & Mexico & PbZnCu & 2 & North America & Sentinel 2 \\
68 & Waihi & New Zealand & AuAg & 3 & Oceania & Sentinel 2 \\
69 & Yauli & Peru & PbZnCu & 5 & South America & Sentinel 2 \\
70 & Yauricocha & Peru & PbZnCu & 5 & South America & Sentinel 2 \\
72 & Zawar & India & PbZn & 2 & Asia & Sentinel 2 \\
74 & Agbaou & Côte d’Ivoire & Au & 1 & Africa & Landsat 8 \\
76 & Batu\_Hijau & Indonesia & CuAu & 1 & Asia & Landsat 8 \\
78 & Bokoni & South Africa & PGEs & 2 & Africa & Landsat 8 \\
79 & Bolivar & Mexico & CuAu & 3 & North America & Landsat 8 \\
80 & Bonikro & Côte d’Ivoire & AuAg & 1 & Africa & Landsat 8 \\
81 & Boss\_Mining\_Group & Dem. Rep. Congo & CuCo & 3 & Africa & Landsat 8 \\
82 & Cadia\_Group & Australia & AuCu & 3 & Oceania & Landsat 8 \\
85 & Cerro\_Corona & Peru & AuCu & 5 & South America & Landsat 8 \\
88 & Collahuasi & Chile & CuMo & 2 & South America & Landsat 8 \\
89 & Costerfield & Australia & Au & 2 & Oceania & Landsat 8 \\
90 & Cozamin & Mexico & PbZnCu & 2 & North America & Landsat 8 \\
91 & Cracow & Australia & AuAg & 3 & Oceania & Landsat 8 \\
92 & Cuajone & Peru & CuMo & 2 & South America & Landsat 8 \\
94 & De\_Grussa & Australia & CuAu & 2 & Oceania & Landsat 8 \\
98 & Essakane & Burkina Faso & Au & 2 & Africa & Landsat 8 \\
100 & Fosterville & Australia & Au & 2 & Oceania & Landsat 8 \\
101 & Franke & Chile & Cu & 2 & South America & Landsat 8 \\
103 & Guelb\_Moghrein & Mauritania & CuAu & 2 & Africa & Landsat 8 \\
104 & Hidden\_Valley & Papua New Guinea & AuAg & 3 & Oceania & Landsat 8 \\
105 & Highland\_Valley & Canada & CuMo & 4 & North America & Landsat 8 \\
106 & Huckleberry & Canada & CuAu & 4 & North America & Landsat 8 \\
108 & Kanmantoo & Australia & CuAu & 2 & Oceania & Landsat 8 \\
109 & Kansanshi & Zambia & CuAu & 3 & Africa & Landsat 8 \\
110 & Khetri\_Group & India & Cu & 2 & Asia & Landsat 8 \\
111 & Kittila & Finland & AuAg & 4 & Europe & Landsat 8 \\
114 & Lady\_Annie & Australia & Cu & 2 & Oceania & Landsat 8 \\
115 & Lagunas\_Norte & Peru & Au & 5 & South America & Landsat 8 \\
116 & Lisheen & Ireland & PbZn & 3 & Europe & Landsat 8 \\
118 & Macraes & New Zealand & Au & 3 & Oceania & Landsat 8 \\
119 & Marlin & Guatemala & AuAg & 3 & North America & Landsat 8 \\
121 & Mission & USA & Cu & 2 & North America & Landsat 8 \\
122 & Mount\_Carlton & Australia & AuCu & 1 & Oceania & Landsat 8 \\
123 & Mount\_Garnet\_Surveyor & Australia & PbZnCu & 3 & Oceania & Landsat 8 \\
124 & Mount\_Milligan & Canada & CuAu & 4 & North America & Landsat 8 \\
126 & Myra\_Falls & Canada & PbZnCu & 3 & North America & Landsat 8 \\
127 & Olimpiada & Russia & AuAg & 4 & Europe & Landsat 8 \\
129 & Oyu\_Tolgoi & Mongolia & CuAu & 2 & Asia & Landsat 8 \\
130 & Parkers\_Mineral\_Hill & Australia & PbZnCu & 2 & Oceania & Landsat 8 \\
131 & Peak & Australia & AuCu & 2 & Oceania & Landsat 8 \\
132 & Pierina & Peru & Au & 5 & South America & Landsat 8 \\
133 & Pilanesberg & South Africa & PGEs & 2 & Africa & Landsat 8 \\
134 & Quebrada\_Blanca & Chile & Cu & 2 & South America & Landsat 8 \\
136 & Ranger & Australia & U & 1 & Oceania & Landsat 8 \\
137 & Ray & USA & Cu & 2 & North America & Landsat 8 \\
138 & Rum\_Jungle & Australia & U-Cu & 1 & Oceania & Landsat 8 \\
139 & Salobo & Brazil & CuAu & 1 & South America & Landsat 8 \\
140 & Savannah\_Sally\_Malay & Australia & Ni S & 2 & Oceania & Landsat 8 \\
141 & Sierra\_Gorda & Chile & CuMo & 2 & South America & Landsat 8 \\
142 & Skorpion & Namibia & PbZn & 2 & Africa & Landsat 8 \\
143 & Sossego & Brazil & CuAu & 1 & South America & Landsat 8 \\
145 & Tenke\_Fungurume & Dem. Rep. Congo & CuCo & 3 & Africa & Landsat 8 \\
147 & Togaruci\_Gosowong & Indonesia & AuAg & 1 & Asia & Landsat 8 \\
148 & Toledo\_Carmen\_Lutopan & Philippines & CuAu & 1 & Asia & Landsat 8 \\
149 & Toquepala & Peru & CuMo & 2 & South America & Landsat 8 \\
150 & Tres\_Valles & Chile & Cu & 2 & South America & Landsat 8 \\
151 & Trident\_Sentinel & Zambia & Cu & 1 & Africa & Landsat 8 \\
152 & Venetia & South Africa & Diamonds & 2 & Africa & Landsat 8 \\
153 & Wetar & Indonesia & Cu & 1 & Asia & Landsat 8 \\
154 & Akyem & Ghana & Au & 1 & Africa & Landsat 8 \\
156 & Broken\_Hill & Australia & PbZn & 2 & Oceania & Landsat 8 \\
157 & Century & Australia & PbZn & 2 & Oceania & Landsat 8 \\
158 & Challenger & Australia & Au & 2 & Oceania & Landsat 8 \\
160 & Cobar\_CSA & Australia & CuAg & 2 & Oceania & Landsat 8 \\
161 & Ernest\_Henry & Australia & CuAu & 2 & Oceania & Landsat 8 \\
164 & Henderson & USA & Mo & 4 & North America & Landsat 8 \\
165 & Kalgoorlie\_SuperPit & Australia & Au & 2 & Oceania & Landsat 8 \\
166 & Kevitsa & Finland & Ni S & 4 & Europe & Landsat 8 \\
169 & La\_Caridad & Mexico & CuMo & 2 & North America & Landsat 8 \\
171 & Lac\_des\_Iles & Canada & PGEs & 4 & North America & Landsat 8 \\
172 & Langlois & Canada & PbZnCu & 4 & North America & Landsat 8 \\
173 & Las\_Cruces & Spain & Cu & 3 & Europe & Landsat 8 \\
175 & Madero & Mexico & PbZnCu & 2 & North America & Landsat 8 \\
176 & Malanjkhand & India & Cu & 1 & Asia & Landsat 8 \\
177 & Mary\_Kathleen & Australia & U & 2 & Oceania & Landsat 8 \\
178 & Masbate & Indonesia & AuAg & 1 & Asia & Landsat 8 \\
179 & Miaogou\_Sanguikou & China & PbZnCu & 2 & Asia & Landsat 8 \\
180 & Mount\_Lyell & Australia & CuAu & 3 & Oceania & Landsat 8 \\
182 & Nifty & Australia & Cu & 2 & Oceania & Landsat 8 \\
183 & Northparkes & Australia & CuAu & 2 & Oceania & Landsat 8 \\
184 & Olympic\_Dam & Australia & CuAu & 2 & Oceania & Landsat 8 \\
185 & Orlovsky & Kazakhstan & PbZnCu & 4 & Asia & Landsat 8 \\
186 & Prominent\_Hill & Australia & CuAu & 2 & Oceania & Landsat 8 \\
187 & Pyhasalmi & Finland & PbZnCu & 4 & Europe & Landsat 8 \\
189 & Reefton & New Zealand & Au & 3 & Oceania & Landsat 8 \\
191 & Rosebery & Australia & PbZnCu & 3 & Oceania & Landsat 8 \\
193 & Tarkwa & Ghana & Au & 1 & Africa & Landsat 8 \\
194 & Telfer & Australia & AuCu & 2 & Oceania & Landsat 8 \\
195 & Tritton & Australia & Cu & 2 & Oceania & Landsat 8 \\
196 & Twangiza & Congo-DRC & Au & 3 & Africa & Landsat 8 \\
198 & Zinkgruvan & Sweden & PbZnCu & 4 & Europe & Landsat 8 \\
199 & Bald\_Mountain & USA & Au & 4 & North America & Landsat 8 \\
201 & Chapada & Brazil & CuAu & 1 & South America & Landsat 8 \\
204 & El\_Salvador & Chile & CuMo & 2 & South America & Landsat 8 \\
205 & Galena & USA & PbZnCu & 4 & North America & Landsat 8 \\
206 & Ghaghoo & Botswana & Diamonds & 2 & Africa & Landsat 8 \\
207 & Golden\_Sunlight & USA & Au & 2 & North America & Landsat 8 \\
210 & Ity & Côte d’Ivoire & Au & 1 & Africa & Landsat 8 \\
211 & Letlhakane & Botswana & Diamonds & 2 & Africa & Landsat 8 \\
212 & Lihir & Papua New Guinea & AuAg & 1 & Oceania & Landsat 8 \\
214 & Marigold\_JV & USA & Au & 2 & North America & Landsat 8 \\
215 & Mount\_Polley & Canada & CuAu & 4 & North America & Landsat 8 \\
217 & Onca\_Puma & Brazil & Ni Lat & 1 & South America & Landsat 8 \\
218 & Pajingo & Australia & AuAg & 2 & Oceania & Landsat 8 \\
219 & Phoenix & USA & AuCu & 2 & North America & Landsat 8 \\
220 & Phu\_Kham & Laos & CuAu & 1 & Asia & Landsat 8 \\
222 & Sabodala & Senegal & Au & 1 & Africa & Landsat 8 \\
224 & Thompson & Canada & Ni S & 4 & North America & Landsat 8 \\
225 & Unki & Zimbabwe & PGEs & 3 & Africa & Landsat 8 \\
226 & Veladero & Argentina & Au & 5 & South America & Landsat 8 \\
\end{longtable}
\end{center}

\newpage
\vspace{1em}
\captionsetup{type=table}
\begin{center}
\begin{longtable}{@{}lllllll@{}}
\label{tab:validation_set} \\
\toprule
\textbf{Mine ID} & \textbf{Mine Name} & \textbf{Country} & \textbf{Commodity} & \textbf{Climate} & \textbf{Region} & \textbf{Sensor} \\
\midrule
\endfirsthead

\toprule
\textbf{Mine ID} & \textbf{Mine Name} & \textbf{Country} & \textbf{Commodity} & \textbf{Climate} & \textbf{Region} & \textbf{Sensor} \\
\midrule
\endhead

1 & Afton\_New\_Afton & Canada & CuAu & 2 & North America & Sentinel \\
2 & Ahafo & Ghana & Au & 1 & Africa & Sentinel \\
7 & Black\_Mountain & Mexico & PbZnCu & 1 & North America & Sentinel \\
11 & Campo\_Morado & Peru & PbZnCu & 1 & South America & Sentinel \\
12 & Candelaria & Chile & Cu & 2 & South America & Sentinel \\
13 & Cannington & Australia & PbZn & 2 & Oceania & Sentinel \\
16 & Cerro\_Corona & Peru & PbZnCu & 5 & South America & Sentinel \\
20 & Chatree & Thailand & Au & 1 & Asia & Sentinel \\
21 & Climax & USA & Mo & 4 & North America & Sentinel \\
22 & Constancia & Peru & CuAu & 5 & South America & Sentinel \\
23 & Contonga & Peru & PbZnCu & 5 & South America & Sentinel \\
25 & El\_Porvenir & Peru & PbZnCu & 5 & South America & Sentinel \\
26 & El\_Soldado & Chile & Cu & 5 & South America & Sentinel \\
27 & El\_Teniente & Chile & CuMo & 5 & South America & Sentinel \\
28 & Erdenet & Mongolia & CuMo & 4 & Asia & Sentinel \\
35 & Kinsevere & DRC & Cu & 3 & Africa & Landsat \\
42 & Modikwa & South Africa & PGEs & 3 & Africa & Sentinel \\
48 & Padcal & Philippines & CuAu & 1 & Asia & Landsat \\
52 & Polska\_Miedz & Poland & CuAg & 4 & Europe & Landsat \\
53 & Priargunsky & Russia & U & 4 & Europe & Landsat \\
56 & Sepon & Laos & CuAu & 1 & Asia & Landsat \\
61 & Thompson & Canada & NiS & 4 & North America & Landsat \\
\bottomrule
\caption{Details of the Validation Set (22 images).} 
\end{longtable}
\end{center}
\vspace{1em}

\newpage

\begin{table}[htbp]
\centering
\label{tab:test_set}
\begin{tabular}{@{}llllll@{}}
\toprule
\textbf{Mine Name} & \textbf{Country} & \textbf{Commodity} & \textbf{Climate} & \textbf{Region} & \textbf{Sensor} \\
\midrule
Bulyanhulu & Tanzania & Au & 1 & Africa & Sentinel 2 \\
Cosala\_Nuestra & Mexico & PbZnCu & 1 & North America & Sentinel 2 \\
Chirano & Ghana & Au & 1 & Africa & Landsat 8 \\
Antucoya & Chile & Au & 3 & South America & Sentinel 2 \\
Goro & New Caledonia & Ni Lat & 1 & Oceania & Landsat 8 \\
Pueblo\_Viejo & Dom. Rep. & Au & 1 & North America & Sentinel 2 \\
Tongon & Côte d'Ivoire & Au & 1 & Africa & Sentinel 2 \\
Santa\_Rita & Brazil & Ni S & 2 & South America & Landsat 8 \\
Orapa & Botswana & Cu & 2 & Africa & Sentinel 2 \\
Tati & Botswana & Ni S & 2 & Africa & Landsat 8 \\
Marula & South Africa & PGEs & 3 & Africa & Sentinel 2 \\
Mount\_Isa & Australia & PbZnCu & 2 & Oceania & Landsat 8 \\
Gabriela\_Mistral & Chile & CuMo & 2 & South America & Landsat 8 \\
Centinela & Chile & Cu & 1 & South America & Sentinel 2 \\
Milpillas & Mexico & Cu & 1 & North America & Landsat 8 \\
Damtshaa & Botswana & Diamonds & 2 & Africa & Landsat 8 \\
Spence & Chile & Cu & 2 & South America & Landsat 8 \\
Antucoya & Chile & Cu & 2 & South America & Sentinel 2 \\
Ravensthorpe & Australia & Ni Lat & 1 & Oceania & Landsat 8 \\
Mantos\_Blancos & Chile & Cu & 2 & South America & Landsat 8 \\
Orapa & Botswana & Diamonds & 2 & Africa & Landsat 8 \\
Granny\_Smith & Australia & Au & 2 & Oceania & Landsat 8 \\
Goldstrike & USA & Au & 2 & North America & Landsat 8 \\
Cerro\_Verde & Peru & Cu & 2 & South America & Sentinel 2 \\
Endeavour & Australia & PbZn & 2 & Oceania & Landsat 8 \\
Cerro\_Color & Chile & Cu & 2 & South America & Landsat 8 \\
Silver\_Bell & USA & Cu & 3 & North America & Sentinel 2 \\
Robinson & USA & CuMo & 2 & North America & Landsat 8 \\
Jwaneng & Botswana & Diamonds & 1 & Africa & Landsat 8 \\
Argyle & Australia & Diamonds & 1 & Oceania & Landsat 8 \\
Konkola\_Nchanga & Zambia & Cu & 1 & Africa & Landsat 8 \\
Bafokeng\_Rasimone & South Africa & PGEs & 2 & Africa & Sentinel 2 \\
Ejimshan & China & Cu & 3 & Asia & Landsat 8 \\
Shishen & China & Cu & 3 & Asia & Landsat 8 \\
Kamoa & DRC & Ni S & 1 & Africa & Landsat 8 \\
Voorspoed & South Africa & Diamonds & 3 & Africa & Landsat 8 \\
Mount\_Rawdon & Australia & AuAg & 3 & Oceania & Landsat 8 \\
Boddington & Australia & AuCu & 3 & Oceania & Landsat 8 \\
Frontier & DRC & Au & 3 & Africa & Sentinel 2 \\
Bingham\_Canyon & USA & CuMoAu & 4 & North America & Landsat 8 \\
Los\_Pelambres & Chile & CuMo & 5 & South America & Landsat 8 \\
Antamina & Peru & PbZnCu & 5 & South America & Sentinel 2 \\
Porgera & Papua New Guinea & AuAg & 3 & Oceania & Landsat 8 \\
Santa\_Rita & Brazil & Ni S & 1 & South America & Landsat 8 \\
\bottomrule
\end{tabular}%
\caption{Details of the Test Set (44 images).}
\end{table}

\begin{table*}[t]
\centering
\label{tab:input_size_patching}
\begin{tabular}{llccccc}
\toprule
\textbf{Dataset} & \textbf{Input Size} & \textbf{Phase} & \textbf{F1 Score} & \textbf{Mean IoU} & \textbf{Accuracy} \\
\midrule
\multirow{3}{*}{FineDataset} & \multirow{3}{*}{128} 
 & Training   & 0.8854 & 0.7451 & 0.9501 \\
 &  & Validation & 0.6878 & 0.5391 & 0.8823 \\
 &  & Testing    & 0.6808 & 0.5459 & 0.8692 \\
\midrule
\multirow{3}{*}{FineDataset + Patching} & \multirow{3}{*}{128}
 & Training   & 0.9435 & 0.7832 & 0.9764 \\
 &  & Validation & 0.6732 & 0.6324 & 0.9244 \\
 &  & Testing    & 0.5373 & 0.4315 & 0.8447 \\
\midrule
\multirow{3}{*}{FineDataset} & \multirow{3}{*}{256}
 & Training   & 0.9333 & 0.8444 & 0.9715 \\
 &  & Validation & 0.7574 & 0.6193 & 0.9034 \\
 &  & Testing    & 0.7539 & 0.6235 & 0.8900 \\
\midrule
\multirow{3}{*}{FineDataset} & \multirow{3}{*}{512}
 & Training   & 0.9290 & 0.8448 & 0.9709 \\
 &  & Validation & 0.7547 & 0.6240 & 0.9121 \\
 &  & Testing    & \textbf{0.7596} & \textbf{0.6391} & \textbf{0.9025} \\
\midrule
\multirow{3}{*}{CoarseDataset} & \multirow{3}{*}{128}
 & Training   & 0.9475 & 0.8913 & 0.9576 \\
 &  & Validation & 0.8411 & 0.7359 & 0.8779 \\
 &  & Testing    & 0.8358 & 0.7285 & 0.8778 \\
\midrule
\multirow{3}{*}{CoarseDataset + Patching} & \multirow{3}{*}{128}
 & Training   & 0.8399 & 0.6354 & 0.8662 \\
 &  & Validation & 0.6628 & 0.5884 & 0.8153 \\
 &  & Testing    & 0.6366 & 0.5089 & 0.7241 \\
\midrule
\multirow{3}{*}{CoarseDataset} & \multirow{3}{*}{256}
 & Training   & 0.9071 & 0.8205 & 0.9250 \\
 &  & Validation & 0.8477 & 0.7450 & 0.8804 \\
 &  & Testing    & \textbf{0.8438} & \textbf{0.7402} & \textbf{0.8811} \\
\midrule
\multirow{3}{*}{CoarseDataset} & \multirow{3}{*}{512}
 & Training   & 0.8710 & 0.7663 & 0.8972 \\
 &  & Validation & 0.8380 & 0.7337 & 0.8760 \\
 &  & Testing    & 0.8336 & 0.7276 & 0.8756 \\
\bottomrule
\end{tabular}
\caption{Performance comparison for different input sizes and the effect of patching on both Fine and Coarse datasets. Each model was evaluated on its corresponding dataset (FineDataset on the Fine test set and CoarseDataset on the Coarse test set).}
\end{table*}

\begin{table}[t]
\centering
\label{tab:fine_test_performance}
\begin{tabular}{lcccccc}
\toprule
\textbf{Dataset} &
\textbf{Total Images} &
\textbf{Min Image Size} &
\textbf{F1 Score} &
\textbf{Mean IoU} &
\textbf{Accuracy} \\
\midrule
Full Coarse                 & 44{,}319 & $7 \times 7$     & 0.5599 & 0.4231 & 0.7060 \\
Coarse-128                  & 12{,}155 & $128 \times 128$ & 0.5841 & 0.4443 & 0.7298 \\
Coarse-FineFiltered         & 1{,}380  & $374 \times 566$ & \textbf{0.5931} & \textbf{0.4552} & \textbf{0.7334} \\
Coarse-FineFiltered-100     & 1{,}280  & $669 \times 362$ & 0.5727 & 0.4335 & 0.7167 \\
\bottomrule
\end{tabular}
\caption{Performance on the fine dataset test split after pre-training on coarse datasets of varying sizes and compositions.}
\end{table}

\twocolumn

\subsection{Fine Datasets Details}

The fine dataset comprises a total of 219 images, which are
split into training (153 images), validation (22 images), and
test (44 images) sets. To ensure a robust evaluation across
diverse global conditions, the dataset is sourced from 39
countries across 6 continents: North America, South Amer-
ica, Africa, Asia, Europe, and Oceania. The dataset is also
balanced across five major climate zones, grouped accord-
ing to the Köppen-Geiger classification system: (1) Tropical,
(2) Arid, (3) Temperate, (4) Cold, and (5) Polar. This com-
prehensive distribution ensures the model is tested against a
wide variety of geographical and environmental backdrops.

\paragraph{Training Set} The training set consists of 153 images, covering a diverse
range of mine types, climates, and geographic locations.
\paragraph{Validation Set} The validation set contains 22 images, used for tuning model
hyperparameters during training.
\paragraph{Test Set} The test set comprises 44 images, held out to provide an
unbiased evaluation of the final model’s performance.

\subsection{Supplementary Experiments}
To determine the optimal hyperparameters and dataset con-
figurations for our pipeline, we conducted a series of abla-
tion studies. These experiments analyze the impact of input
image resolution, the use of patching, and the composition
of the coarse dataset on model performance.
Figure 1 visually corroborates our findings and highlights
the limitations of other approaches. As shown, the outputs
from standard segmentation architectures (c-e) and other
continual learning strategies (f-g) struggle with the task’s
complexity, producing incomplete masks that fail to capture
fine, connecting structures. In contrast, the predictions from
our proposed method, MiningNet (h) are qualitatively su-
perior, aligning almost perfectly with the ground truth (b)
by accurately delineating complex boundaries to generate a
complete and precise segmentation mask.

\paragraph{Continual Learning Strategy} To validate our frame-
work’s core design, we compare it against established Con-
tinual Learning (CL) strategies. For our coarse-to-fine task,
these methods are re-contextualized: the ”continual” pro-
cess involves transferring knowledge from a teacher model
(trained on coarse data) to a student model (trained on fine
data). We adapted LwF \cite{Zhizhong2018}, which uses
knowledge distillation (KD) as a regularizer to preserve the
teacher’s general knowledge, and LwM \cite{dhar2019},
which in our case uses attention distillation to focus the stu-
dent model on critical boundary regions. Finally, we adopt
a selective Replay strategy \cite{rolnick2019} inspired
by experience replay in reinforcement learning. In our ap-
proach, we filter coarse-labeled samples based on a prede-
fined quality heuristic: a model trained on fine data is used
to generate predictions on the coarse images, and only sam-
ples with high consistency between the prediction and the
coarse label are retained for joint training. While these CL
adaptations offer benefits, our proposed MiningNet still sur-
passes them, demonstrating the advantage of its advanced
attentive distillation mechanism for more effective and tar-
geted knowledge transfer.

\paragraph{Impact of Input Resolution and Patching} We first evaluated the effect of different input sizes (128 ×
128, 256 × 256, and 512 × 512 pixels) on both the Fine
and Coarse datasets. Additionally, we investigated whether
an image patching strategy could improve performance. For
this experiment, we utilized the patchify Python library,
a tool designed to split images into a grid of smaller, po-
tentially overlapping patches. We configured it to extract
patches of 128 × 128 pixels with a step size of 128 to
avoid overlap between adjacent patches. The hypothesis was
that training on these patches would allow the model to learn more fine-grained local features from the original high-
resolution imagery, which is otherwise lost when down-
scaling the entire image. By processing these high-detail
patches, we aimed to improve the model’s ability to segment
intricate boundaries. The results for all experiments are de-
tailed in Table 4.

From the results, it is evident that an input size of 256 ×
256 or 512 × 512 provides the best performance for the fine-
tuned model, with the 512 × 512 input achieving a slightly
higher testing F1-score (0.7596 vs 0.7539). For the coarse
dataset, the 256 × 256 input size clearly outperforms the
others, achieving the highest testing F1-score of 0.8438.
Notably, the application of patching to the 128 × 128 in-
put proved to be detrimental, significantly degrading perfor-
mance for both datasets. This suggests that the model bene-
fits more from a complete view of the image, even at a lower
resolution, than from localized patches. Given the trade-off
between performance and computational cost, an input size
of 256×256 was selected as the optimal choice for our main
experiments.

\paragraph{Impact of Coarse Dataset Composition} The coarse dataset serves as a large-scale pre-training cor-
pus. Our initial set contained 44,319 images with extreme
variations in dimension, from as small as 7x7 pixels to
over 1300x1400 pixels. A significant portion of this dataset
(32,164 images) consisted of very low-resolution images
smaller than 128x128 pixels. To find the most effective pre-
training data, we tested several filtering strategies based on
minimum image dimensions. We evaluated the full, unfil-
tered dataset; a version filtered to include only images of
at least 128x128 pixels (‘Coarse-128‘); and a more targeted
version (‘Coarse-FineFiltered‘) designed to align the coarse
data’s characteristics with our fine dataset, using the small-
est fine mine’s area (9.81 km²) as a benchmark. The perfor-
mance of a model pre-trained on each of these coarse dataset
versions was then evaluated on the fine dataset’s test split.
The results are shown in Table 5.

The results demonstrate the critical importance of curat-
ing the coarse dataset. Using the full, unfiltered set of 44,319
images yields the poorest performance, likely due to the
noise introduced by the thousands of very low-resolution
images. Filtering the data to remove images smaller than
128x128 pixels (‘Coarse-128‘) provides a notable improve-
ment. However, the best results are achieved with the
Coarse-FineFiltered set. This approach, which curated the
coarse data to align with the area characteristics of the fine
dataset, produced the highest F1-score of 0.5931. This in-
dicates that pre-training on a well-curated, domain-relevant
dataset is more effective than using a larger, noisier one.

\subsection{Training}
In this sub-section, we describe how to train the model, including data augmentation, model configuration, optimization parameters, evaluation metrics, and baselines.

\subsubsection{Data Augmentation}
To enhance the model's robustness and generalizability, we applied a series of data augmentations to the training set. These included random horizontal and vertical flips, random 90-degree rotations, and a combination of shifting, scaling, and rotation with a scale limit of 0.3, a rotation limit of 15 degrees, and a shift limit of 0.1. Each augmentation was applied with a probability of 0.5.

 \subsubsection{Model Configuration and Implementation}
Theoretically, our CL approach can be applied to any deep learning models. However, in our implementation, a Feature Pyramid Network (FPN) with a DenseNet-121 backbone, pre-trained on ImageNet, were used for both the teacher and the student models. The models were configured to accept 6-channel inputs (RGB, NIR, and SWIR) with a spatial dimension of $256 \times 256$ pixels. For the student model, which incorporates features from both coarse and fine data, the entire network was fine-tuned during training. During the 2nd stage, the teacher model was frozen to act as a stable knowledge source. All models were implemented using Python 3.9 and TensorFlow 2.13. The training and evaluation were conducted on a PC workstation with an Intel Xeon W-1350 CPU and an NVIDIA RTX A4000 GPU with 16GB of VRAM.

\subsubsection{Training Parameters and Optimization}
We employed the AdamW \cite{loshchilov2019} optimizer, a variant of Adam that decouples weight decay from the gradient updates, with an initial learning rate of $1 \times 10^{-3}$ and a weight decay of $1 \times 10^{-4}$. To manage the learning rate dynamically, we implemented a custom schedule that begins with a warm-up phase for the first 5 epochs, linearly increasing the learning rate to its base value. Following the warm-up, a \texttt{ReduceLROnPlateau} callback was activated, which reduces the learning rate by 0.5 if the validation Mean IoU (\texttt{val\_Mean\_IoU}) did not improve for 3 consecutive epochs. The models were trained for a maximum of 100 epochs with a batch size of 8. To prevent overfitting and ensure the retention of the best-performing model, we utilized an \texttt{EarlyStopping} callback that monitored the \texttt{val\_Mean\_IoU} with a patience of 10 epochs. The model weights that yielded the highest \texttt{val\_Mean\_IoU} on the validation set were saved and used for the final evaluation.

\subsubsection{Baselines and Evaluation Metrics}
We compare MineC2FNet against 2 groups of baselines: 1) CNN and Transformer-based models, and 2) CL-based methods. For the CNN and Transformer-based models, we trained FPN, DeepLabV3+, U-Net, and Prithvi (IBM-NASA geospatial foundation model) directly on our fine-grained dataset to establish a strong performance. For the CL-based methods, since no prior work addresses our coarse-to-fine CL domain adaptation task, we adapt LwF \cite{Zhizhong2018}, LwM \cite{dhar2019}, and Replay strategy \cite{rolnick2019} such that the training involves transferring knowledge from a teacher model (trained on coarse data) to a student model (trained on fine data). All models were trained using RGB, NIR, and SWIR bands as the input. Finally, we evaluate our framework and the baselines using standard semantic segmentation metrics: \textit{Accuracy}, \textit{Mean F1 Score} (for class imbalance), and \textit{Mean Intersection over Union (mIoU)}.

\subsection{Results and Discussion}

\paragraph{Performance across Köppen-Geiger climate zones}To assess the model robustness and the generalizability of its domain adaptation, we evaluated the model across five Köppen–Geiger climate zones \cite{Beck2018}, covering a wide range of global climates, from tropical to polar. The results in Table \ref{tab:climate_performance} confirm MineC2FNet adaptability as it performs exceptionally well in arid and tropical zones, where mining activities are often concentrated, and achieves a peak mIoU of 77.34\% in temperate climates. The performance in polar (69.42\%) and cold (57.47\%) zones, even with very limited data, proves that our domain-adapted model can effectively differentiate mining footprints across diverse geographic locations, including in underrepresented climate types and homogenous terrains.

\begin{table}[t]
\centering
\small
\setlength{\tabcolsep}{1mm}
\begin{tabular}{lccc}
\toprule
\textbf{Climate Zone (train/val/test)} & \textbf{Acc.} & \textbf{mF1} & \textbf{mIoU} \\
\midrule
Tropical (25/4/8)   & 95.19 & 86.44 & 67.63 \\
Arid (63/4/21)      & 90.78 & 82.93 & 72.21 \\
Temperate (24/4/9)  & \textbf{95.24} & \textbf{86.45} & \textbf{77.34} \\
Cold (16/4/1)       & 85.82 & 72.99 & 57.47 \\
Polar (11/4/2)      & 87.41 & 81.87 & 69.42 \\
\bottomrule
\end{tabular}
\caption{Performance across Köppen-Geiger climate zones.}
\label{tab:climate_performance}
\end{table}

\begin{table}[t]
\centering
\small
\setlength{\tabcolsep}{1mm} 
\begin{tabular}{lcc}
\toprule
\textbf{Metric} & \textbf{FPN} & \textbf{MineC2FNet (Ours)} \\
\midrule
mIoU            & 0.6825 & \textbf{0.7364 (+5.4)} \\
Avg. Time (s)   & 0.0748 & 0.0909 (+0.0161) \\
Speed (FPS)     & 13.38  & 11.00 (-2.38) \\
\bottomrule
\end{tabular}
\caption{Model complexity and performance comparison between the baseline FPN and the proposed MineC2FNet.}
\label{tab:model_complexity}
\end{table}

\subsubsection{Model Complexity and Efficiency}
Finally, we evaluate the computational cost of our proposed MineC2FNet against the FPN baseline, with results presented in Table \ref{tab:model_complexity}. The table clearly shows that MineC2FNet achieves a substantial performance gain, improving the mIoU by 5.4 points over the baseline.

This significant improvement in accuracy comes with a moderate trade-off in computational efficiency. Specifically, our model's average inference time increased by 0.0161 seconds, resulting in a processing speed of 11.00 FPS compared to the baseline's 13.38 FPS. This decrease in speed is an expected consequence of incorporating the more sophisticated selective continual learning mechanism (attentive feature injection and knowledge transfer). We argue that the considerable gain in segmentation accuracy justifies this modest increase in computational overhead for our remote sensing application, where model performance is often the primary consideration.

\begin{table}[t] 
\centering
\resizebox{\columnwidth}{!}{%
\begin{tabular}{lcccc}
\hline
\textbf{Configuration} & \textbf{Acc.(\%)} & \textbf{F1(\%)} & \textbf{mIoU(\%)} & \textbf{$\Delta$mIoU} \\ \hline
BCE & 0.9096 & 0.7733 & 0.6518 & - \\
BCE+DICE & 0.9016 & 0.7783 & 0.6545 & +0.0027 \\
BCE+DICE+SSIM & 0.9039 & 0.7832 & 0.6642 & +0.0124 \\
\textbf{BCE+DICE+SSIM+Bnd.} & \textbf{0.9102} & \textbf{0.7874} & \textbf{0.6687} & \textbf{+0.0169} \\ \hline
\end{tabular}%
}
\caption{Ablation Study on Student Loss Components}
\label{tab:loss_ablation}
\end{table}

\subsubsection{Impact of Student Loss Combinations}
We conducted an ablation study to determine the optimal configuration for the student loss function, with a specific focus on enhancing edge delineation (see Table \ref{tab:loss_ablation}). While the baseline Binary Cross-Entropy (BCE) provides a foundation for pixel-wise classification, it struggles with structural integrity in fine-grained mining masks. The incremental addition of Dice and SSIM losses improved regional overlap and geometric shape preservation, respectively. Notably, the final inclusion of Boundary Loss yielded the highest performance (0.6687 mIoU) by weighting optimization toward edge pixels. This four-part combination effectively eliminates segmentation "blurring," ensuring the student model achieves the most precise delineation of mining site perimeters.

\begin{table}[t]
\centering
\resizebox{\columnwidth}{!}{%
\begin{tabular}{lcccc}
\hline
\textbf{Method} & \textbf{TN} & \textbf{FP} & \textbf{FN} & \textbf{TP (Mining)} \\ \hline
MDIL-SS [Garg et al., 2022] & 67.90 & 4.96 & 6.42 & 20.72 \\
GSMF-RS-DIL [Huang et al., 2024] & 69.63 & 5.89 & 5.28 & 19.20 \\
\textbf{MineC2FNet (Ours)} & \textbf{68.10} & \textbf{3.52} & \textbf{4.15} & \textbf{24.23} \\ \hline
\end{tabular}%
}
\caption{Comparison of Confusion Matrix Results (\%)}
\label{tab:cm_comparison}
\end{table}

\subsubsection{Confusion Matrix analysis}
Based on our analysis of the confusion matrix (see Table \ref{tab:cm_comparison}), we provide quantitative evidence that MineC2FNet prioritizes foreground precision, a factor critical for high-fidelity environmental monitoring. While competing domain-incremental learning (DIL) methods, such as GSMF-RS-DIL, achieve high background classification accuracy, they exhibit higher False Positive Rates (FPR) and lower True Positive Rates (TPR) for the mining class. Specifically, MineC2FNet restricts the False Positive (FP) rate to 3.52\%, marking a reduction of approximately 29–40\% relative to MDIL-SS (4.96\%) and GSMF-RS-DIL (5.89\%). This transition from background-biased classification to foreground-focused precision directly accounts for the +2\% IoU gap (see Table 1 in the main paper); by suppressing erroneous mining detections while simultaneously capturing more authentic mining footprints, MineC2FNet offers a more robust framework for tracking industrial expansion.

\end{document}